\documentclass[10pt,journal,compsoc]{IEEEtran}
\ifCLASSOPTIONcompsoc
  \usepackage[nocompress]{cite}
\else
  \usepackage{cite}
\fi

\ifCLASSINFOpdf
\usepackage[pdftex]{graphicx}
\else
\fi
\usepackage{amsmath}

\usepackage{bm}
\usepackage[dvipsnames]{xcolor}
\usepackage{newtxmath}
\usepackage{multirow}
\usepackage{multicol}
\usepackage{adjustbox}
\usepackage{booktabs}
\usepackage{makecell}
\usepackage{float}
\usepackage{caption}
\usepackage{latexsym}
\usepackage{subcaption}
\usepackage{makecell}
\usepackage{pifont}

\usepackage{fixltx2e}

\usepackage[pagebackref=true,breaklinks=true,colorlinks,bookmarks=false]{hyperref}

\begin{document}
%
%

\title{SPAN: Learning Similarity between Scene Graphs and Images with Transformers}
%
%
%

\author{Yuren Cong, Wentong Liao,  Bodo Rosenhahn, and Michael Ying Yang 
\IEEEcompsocitemizethanks{\IEEEcompsocthanksitem Yuren Cong, Wentong Liao and Bodo Rosenhahn are with Institute of Information Processing, Leibniz University Hannover, Germany.
E-mail: \url{{cong, liao, rosenhahn}@tnt.uni-hannover.de}.
\IEEEcompsocthanksitem Micheal Ying Yang is with Viusal Computing Group, University of Bath, UK. Email:
\url{myy35@bath.ac.uk}.}
}

\IEEEtitleabstractindextext{%
\begin{abstract}
Learning similarity between scene graphs and images aims to estimate a similarity score given a scene graph and an image.
There is currently no research dedicated to this task, although it is critical for scene graph generation and downstream applications.
Scene graph generation is conventionally evaluated by Recall$@K$ and mean Recall$@K$, which measure the ratio of predicted triplets that appear in the human-labeled triplet set.
However, such triplet-oriented metrics fail to demonstrate the overall semantic difference between a scene graph and an image and are sensitive to annotation bias and noise.
Using generated scene graphs in the downstream applications is therefore limited.
To address this issue, for the first time, we propose a \textbf{S}cene gra\textbf{P}h-im\textbf{A}ge co\textbf{N}trastive learning framework, \textbf{SPAN}, that can measure the similarity between scene graphs and images.
Our novel framework consists of a graph Transformer and an image Transformer to align scene graphs and their corresponding images in the shared latent space.
We introduce a novel graph serialization technique that transforms a scene graph into a sequence with structural encodings.
Based on our framework, we propose R-Precision measuring image retrieval accuracy as a new evaluation metric for scene graph generation.
We establish new benchmarks on the Visual Genome and Open Images datasets.
Extensive experiments are conducted to verify the effectiveness of SPAN, which shows great potential as a scene graph encoder.
\end{abstract}

}

\maketitle

\IEEEdisplaynontitleabstractindextext

%
\IEEEpeerreviewmaketitle

\IEEEraisesectionheading{\section{Introduction}\label{sec:introduction}}

%
%
%
%

\IEEEPARstart{A}{}scene graph is a graphical representation in which nodes symbolize the entities in a scene and the edges indicate the relationships between the entities~\cite{johnson2015image}.
It is regarded as a compact representation for scene understanding, as well as a promising tool to bridge the domains of vision and language.
However, learning similarity between scene graphs and images is still a gap.
This task is critical for scene graph generation (SGG) and downstream applications of scene graphs, such as image retrieval~\cite{johnson2015image, schroeder2020structured, Wang_2020_WACV, yoon2021image}, image generation~\cite{johnson2018image, li2019pastegan, herzig2020learning, yang2022diffusion}, image captioning~\cite{gao2018image, yang2019auto}.

\begin{figure}[ht!]
\centering
\includegraphics[width=1\linewidth]{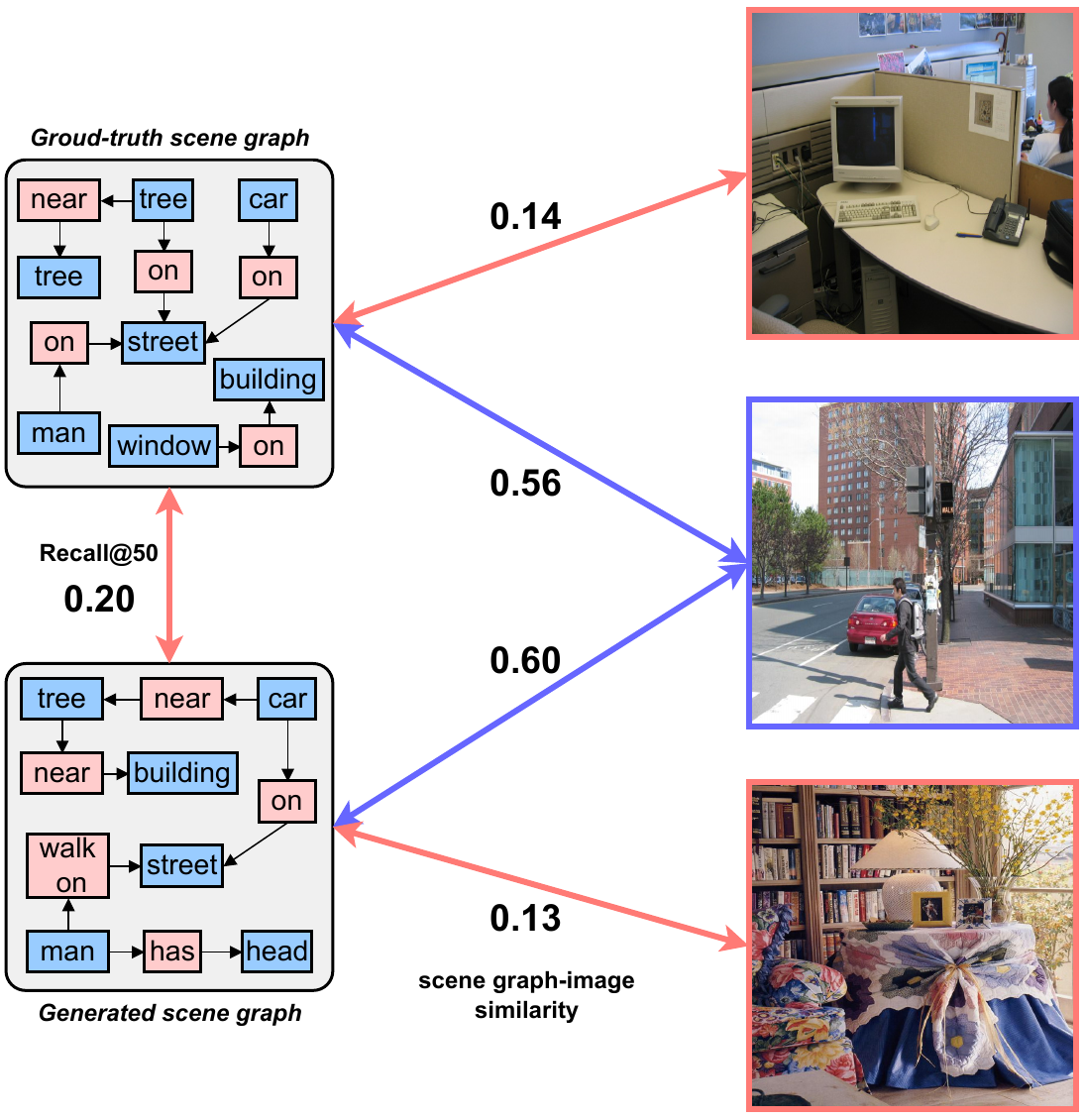}
\captionof{figure}{Our method can quantitatively measure the scene graph-image similarity. 
The ground truth and generated scene graphs consist of different triplets with a low Recall$@50$, but both represent a street scene.
Our method assigns higher similarity scores to the street scene image compared to the irrelevant images. 
}
\label{fig:teaser}
\end{figure}


Learning the similarity between scene graphs and images demonstrates the overall semantic difference between scene graphs and images.
This quantification can facilitate the evaluation of scene graph generation.
Recently, scene graph generation methods are developing rapidly~\cite{zellers2018neural,yan2020pcpl,  li2021bipartite, tang2020unbiased, chiou2021recovering, yang2017support}, while the issues in evaluating the quality of the generated scene graphs are largely overlooked.
All existing SGG works adopt Recall$@K$ and mean Recall$@K$ as their evaluation metrics, which measure the ratio of predicted triplets (\texttt{$<$subject-predicate-object$>$}) that appear in the human-labeled triplets.
However, when the manually labeled triplets are biased or noisy (which is inevitable in scene graph datasets), such evaluation metrics cannot present the real quality of the generated scene graphs.
For example, consider the street scene in Fig.~\ref{fig:teaser},
the generated scene graph from \cite{10105507} has a low Recall$@50=0.20$ while the predicted relationships are correct.
This risk increases when the scene graph scales up.
The triplet-oriented evaluation metrics limit the development of scene graph generation.
Furthermore, the ambiguous quality of the generated scene graph also limits its application in downstream vision-language tasks.
To overcome these issues, there is an urgent demand for a graph-oriented metric to measure how well the overall semantic information of a scene graph matches the visual appearance of a given image.


In this paper, for the first time, we propose 
a \textbf{S}cene gra\textbf{P}h-im\textbf{A}ge co\textbf{N}trastive learning framework, \textbf{SPAN}, to align scene graphs and their corresponding images in the shared latent space~\footnote{The source code is  publicly available at \url{https://github.com/yrcong/Learning_Similarity_between_Graphs_Images}.}.
Our method can quantitatively measure the similarity between scene graphs and images, as depicted in~Fig.~\ref{fig:teaser}.
SPAN consists of a graph Transformer (as a graph encoder) and an image Transformer (as an image encoder), similar as CLIP \cite{radford2021learning}.
Since Transformers cannot process graph data directly, we introduce a novel graph serialization approach that flattens the nodes and edges of the scene graph into a sequence.
We also design specific structural encodings to preserve graph structure information.
Based on our framework, a R-Precision is proposed, which measures image retrieval accuracy as a new evaluation metric for scene graph generation.
Compared with triplet-oriented metrics, our graph-oriented metric is more robust against data bias and noise. 
Our \textbf{main contributions} are summarized as follows:
\begin{itemize}
\item We propose a novel contrastive learning framework based on Transformers to learn the similarity between scene graphs and images.
\item We introduce a graph serialization method that enables our graph Transformer to comprehend the graph structure. It can be used as a generic scene graph encoder.
\item A novel metric R-Precision based on our framework is introduced to evaluate scene graph generation by using the similarity between scene graphs and images.
\item We create new benchmarks on Visual Genome \cite{krishna2017visual} and Open Images \cite{kuznetsova2020open}, and perform a systematic analysis of the existing scene graph generation methods.
\end{itemize}

The remainder of the paper is structured as follows.
In Section~\ref{sec:related_work}, we review related work in scene graph generation. 
Section~\ref{sec:method} presents our proposed scene graph-image contrastive learning framework.
Section~\ref{sec:metrics} reviews the common evaluation metrics for scene graph generation and introduces the new R-Precision metric.
Experimental results of the proposed framework are discussed in Section~\ref{sec:exp}.
Section~\ref{sec:con} concludes this paper.

\section{Related Work}
\label{sec:related_work}

\subsection{Scene graphs}
Scene graphs have been shown to be promising for other vision tasks like image retrieval~\cite{johnson2015image, yoon2021image}, image captioning~\cite{yang2019auto}, image generation~\cite{johnson2018image, li2019pastegan, yang2022diffusion}, and visual question answering~\cite{shi2019explainable, lee2019visual, hildebrandt2020scene}. 
A precise alignment between scene graphs and images is critical for these applications.
For image retrieval, a conditional random field is constructed in~\cite{johnson2015image} to model the distribution over all possible scene graph groundings, and the maximum likelihood of a posterior inference is taken as the score measuring the similarity between the scene graph and the image.
IRSGS~\cite{yoon2021image} uses scene graphs as the medium for image-to-image retrieval by measuring the similarity between scene graphs. 
For image captioning, SGAE~\cite{yang2019auto} learns a dictionary to encode the language inductive bias from scene graphs to natural language sentences. 
For image generation, a graph convolutional network is utilized in~\cite{johnson2018image} to extract semantic information from the scene graph to predict the segmentation masks and the scene layout. 
Predicting the scene layout from scene graphs becomes the main paradigm for scene graph-to-image generation~\cite{herzig2020learning,ashual2019specifying,li2019pastegan}.
The above applications usually use manually labeled scene graphs, which are high cost to annotate.
As a result, a large number of SGG models have emerged~\cite{yang2018graph, gu2019scene, li2022devil, dhingra2021bgt}.
To evaluate the quality of generated scene graphs, Recall$@K$~\cite{lu2016visual} and mean Recall$@K$~\cite{tang2019learning} are widely adopted.
However, Recall$@K$ and mean Recall$@K$ compare the predicted triplets with the ground truth relationships strictly, which makes them easily impacted by noise and bias in the dataset.
Additionally, such triplet-oriented metrics fail to demonstrate the overall semantic difference between a scene graph and an image.
%
%
In this work, we fill this gap by proposing a contrastive learning framework connecting scene graphs and images, which can measure the similarity between scene graphs and images.

\begin{figure*}[t!]
\centering
\includegraphics[width=0.96\linewidth]{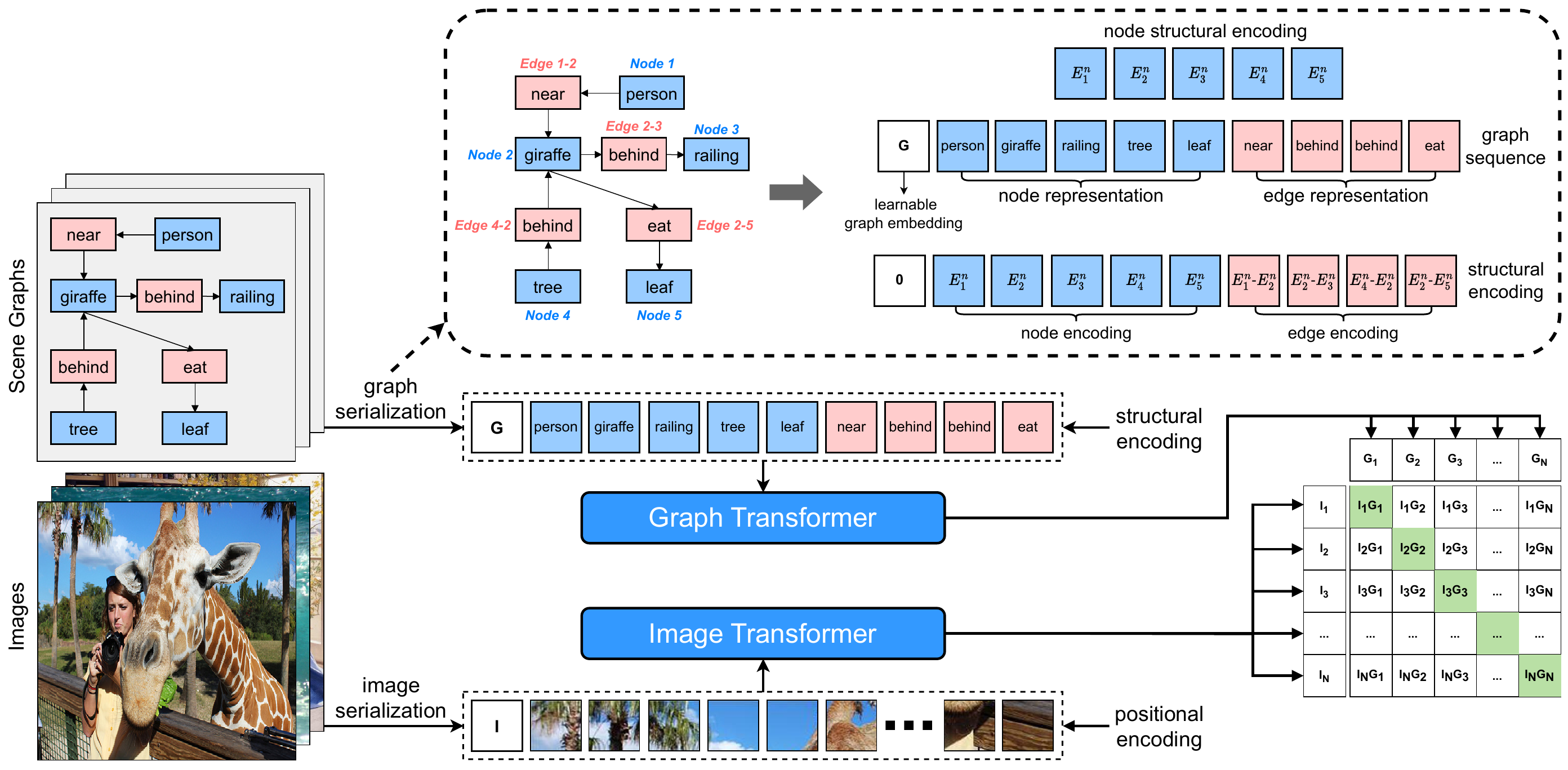}
\caption{Overview of our method: SPAN is a novel contrastive learning framework consisting of a scene graph Transformer and an image Transformer to reason about scene graph and image representations, respectively. We use a contrastive loss to align the scene graph and its corresponding image in the shared latent space. 
In order to enable the Transformer to understand the scene graph data, we propose a graph serialization approach and introduce a graph structural encoding. 
The node representation, edge representation, node encoding, graph embedding $\bm{G}$, and image embedding $\bm{I}$ are all learnable embeddings.}
\label{fig:sicon_framework}
\end{figure*}

\subsection{Multi-modal contrastive learning}
Contrastive learning is a representation learning which aims to group similar samples closer and diverse samples far from each other \cite{jaiswal2020survey}.
It plays an important role in self-supervised learning and can address the limitations of supervised learning.
Virtex~\cite{desai2021virtex} uses textual annotations of images on COCO Captions \cite{chen2015microsoft} to learn visual representations. 
To make use of the myriad available images with textual descriptions, CLIP \cite{radford2021learning} pre-trains a visual-language model with the contrastive loss using 400 million image-text pairs collected from a variety of publicly available sources. The pre-trained image encoder and text encoder have powerful representations and have been successfully applied in many visual-language tasks.
RLIP~\cite{yuan2022rlip,yuan2023rlipv2} establishes correspondences from both entities and relations to free-form text descriptions.
Compare to the rapid progress of visual and language representation learning, the representation learning for scene graphs is far under-explored.
Contrastive learning is firstly used for scene graph generation in \cite{zhang2019graphical} to solve the entity instance confusion and the relationship ambiguity.
A graph convolutional network (GCN) is trained with image-masked auto-encoding loss in \cite{yang2022diffusion}, and then use it for scene graph-to-image generation.
GCNs \cite{kipf2016semi} are popular choices for encoding scene graphs \cite{li2019pastegan, Wang_2020_WACV, herzig2020learning, yang2022diffusion}.
However, GCNs are  designed to operate on local neighborhoods of nodes, which makes them less effective at capturing global representations.
In this work, we propose a contrastive learning framework based on Transformers which is able to extract high quality representations for scene graphs and images.  
To the best of our knowledge, we are the first for this task.

\section{Scene Graph-Image Contrastive Learning}
\label{sec:method}

We propose a novel \textbf{S}cene gra\textbf{P}h-im\textbf{A}ge co\textbf{N}trastive learning framework, \textbf{SPAN}, to measure the similarity between scene graphs and images.
It consists of a scene graph encoder and an image encoder, both of which are built on the Transformer architecture~\cite{vaswani2017attention}. 
The overview of our framework is demonstrated in~Fig.~\ref{fig:sicon_framework}.
The input scene graph can be a location-free scene graph (where nodes only represent entity categories) or a location-bounded scene graph (where nodes represent entity categories and bounding boxes).

\subsection{Scene Graph Representation}


To capture graph representations, graph convolutional networks (GCNs) are conventionally adopted.
However, GCNs are primarily designed for local neighborhoods of nodes while even unconnected nodes in a scene graph can be highly correlated.
Therefore, GCNs are less effective in capturing scene graph representations.
In this paper, we use a Transformer to reason about representative scene graph features.
To adapt the graph-structured data to Transformer, we introduce a novel approach to serialize scene graphs.


Given a scene graph with $M$ entity nodes and $N$ edges for the relationships, a scene graph sequence can be constructed by concatenating node representations $\{\bm{n}_1, \bm{n}_2,...,\bm{n}_M\}$ and edge (relation) representations $\{\bm{r}_1,\bm{r}_2,...,\bm{r}_N\}$.
For location-free scene graphs, the node representation $\bm{n}\in\mathbb{R}^{d}$ is the learnable embedding corresponding to the entity category.
For location-bounded scene graphs, the bounding box coordinates are transformed into box embeddings by linear projection layers.
The box embedding is further element-wise added to the corresponding node representation.

\begin{figure}[t!]
\centering 
\includegraphics[width=0.97\linewidth]{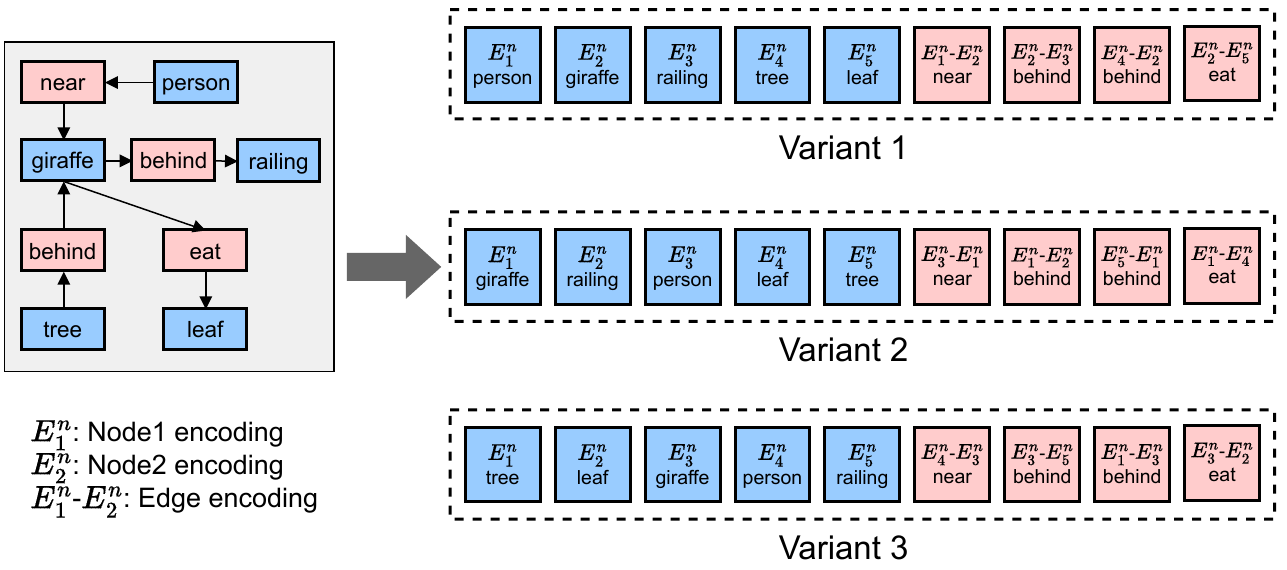}
\vspace{-2mm}
\caption{To avoid potential biases resulting from the consistent assignment of node encodings to specific node representations during training, we randomly shuffle the nodes and assign them different structural encodings during each iteration.}
\label{fig:graph_seq}
\end{figure}

As the serialized scene graph sequence does not characterize the original graph structure information by itself, we design a structural encoding and integrate it into the graph sequence to enable the Transformer to understand the scene graph data. 
We first construct $M$ learnable node structural encodings $\{\bm{E}^n_1, \bm{E}^n_2,...,\bm{E}^n_M\}$ to differentiate the $M$ nodes within a scene graph and utilize the node encodings to determine the edge structural encodings $\{\bm{E}^r_1, \bm{E}^r_2,...,\bm{E}^r_N\}$ (see~Fig.~\ref{fig:sicon_framework}). 
For a relation edge $\bm{r}_k$ pointing from the start node $\bm{n}_i$ to the end node $\bm{n}_j$, its edge encoding can be computed as:
\begin{equation}
\centering
\bm{E}^r_k=\bm{E}^n_i-\bm{E}^n_j,
\end{equation}
where $\bm{E}^n_i$ is the node encoding of $\bm{n}_i$ and $\bm{E}^n_j$ is the node encoding of $\bm{E}^n_j$.
The scene graph sequence $S_G$ and the corresponding structural encoding $E_G$ can be formulated as:
\begin{equation}
\centering
\begin{aligned}
& S_{G} =[\bm{n}_1, \bm{n}_2,...,\bm{n}_M, \bm{r}_1,\bm{r}_2,...,\bm{r}_N] \\
& E_{G} =[\bm{E}^n_1, \bm{E}^n_2,...,\bm{E}^n_M\, \bm{E}^r_1, \bm{E}^r_2,...,\bm{E}^r_N].
\end{aligned}
\label{eq:graph_sequence_encoding}
\end{equation}
When assigning the node encodings to the node representations, the sequence order is implicitly established. This order information is not originally present in the graph structure. 
To avoid potential biases resulting from the consistent assignment of node encodings to specific node representations during training,  we implement a process where nodes are shuffled and assigned different node encodings across various training iterations.
For the same scene graph, the serialization results differ due to node shuffle. 
An example is given in~Fig.~\ref{fig:graph_seq}.
In the three variants, the node encodings $\bm{E}^n_1$, $\bm{E}^n_3$, and $\bm{E}^n_4$ are respectively assigned to the node \texttt{$<$person$>$}.
This approach leads to diverse serialization outcomes for the same graph,  encouraging node encodings to learn robust structural information of scene graphs.

We use the vanilla Transformer encoder stack~\cite{vaswani2017attention} to reason about scene graph representations. 
Each encoder layer consists of a multi-head attention module, a feed-forward network (FFN), and two normalization layers.
Similar to the \texttt{[class]} token used in BERT~\cite{devlin2018bert}, a learnable graph embedding $\bm{G}\in\mathbb{R}^{d}$ for context is inserted into the scene graph sequence to capture the scene graph representation. 
No additional encoding is introduced for this graph embedding $\bm{G}$.
In the multi-head attention module, the graph structural encoding $E_G$ is added to the scene graph sequence $S_G$ for the query and key, while the value is $S_G$ directly. 
The graph embedding $\bm{g}$ output by the last encoder layer is adopted as the final scene graph representation. 

\subsection{Image Representation}
\begin{figure}[tp]
\centering
\includegraphics[width=0.96\linewidth]{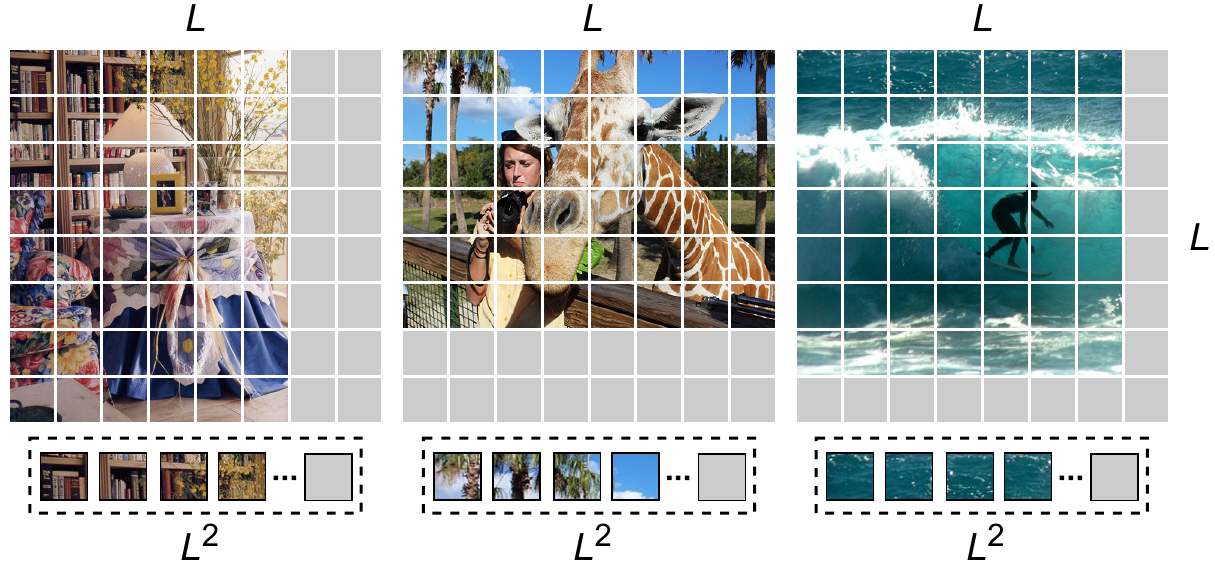}
\caption{In order to maintain consistency between scene graphs and images, input images with different aspect ratios in the batch are zero-padded to a square of length $L$, rather than being cropped and resized.
As a result, the length of resultant sequences is $L^2$. Zero-padded patches are masked off in the image Transformer.
}
\label{fig:image_seq}
\end{figure}

We learn image representations also with a Transformer while keeping the original aspect ratio of the input images.
We use a pre-trained backbone~\cite{he2016deep} to extract the feature map and a convolution layer to transform the channel dimension of the feature map to the model dimension $d$.
For image pre-processing, we do not perform image cropping, which will decrease the consistency between the scene graph and the image. 
To make the batch operation feasible, the images with different sizes in the batch are zero-padded to the same shape of $L\times L \times 3$, as depicted in~Fig.~\ref{fig:image_seq}.
Then, the feature maps are serialized into a feature sequence of length $L^2$.
The features of the zero-padded patches are masked off in the image Transformer.

Similar to the graph encoder, we adopt the vanilla Transformer encoder stack to reason about image representations.
To capture the image representation, we also construct a learnable image embedding $\bm{i}\in\mathbb{R}^{d}$ and incorporate it into the image feature sequence.
As the Transformer architecture is permutation-invariant, fixed positional encodings~\cite{vaswani2017attention, carion2020end} are introduced to facilitate the recognition of the relative positions and spatial layout of the patch features.
The image embedding $\bm{i}$ output by the last encoder layer is used as the final image representation.

\subsection{Transformer Architecture}
\begin{figure}[tp]
\centering
\includegraphics[width=0.95\linewidth]{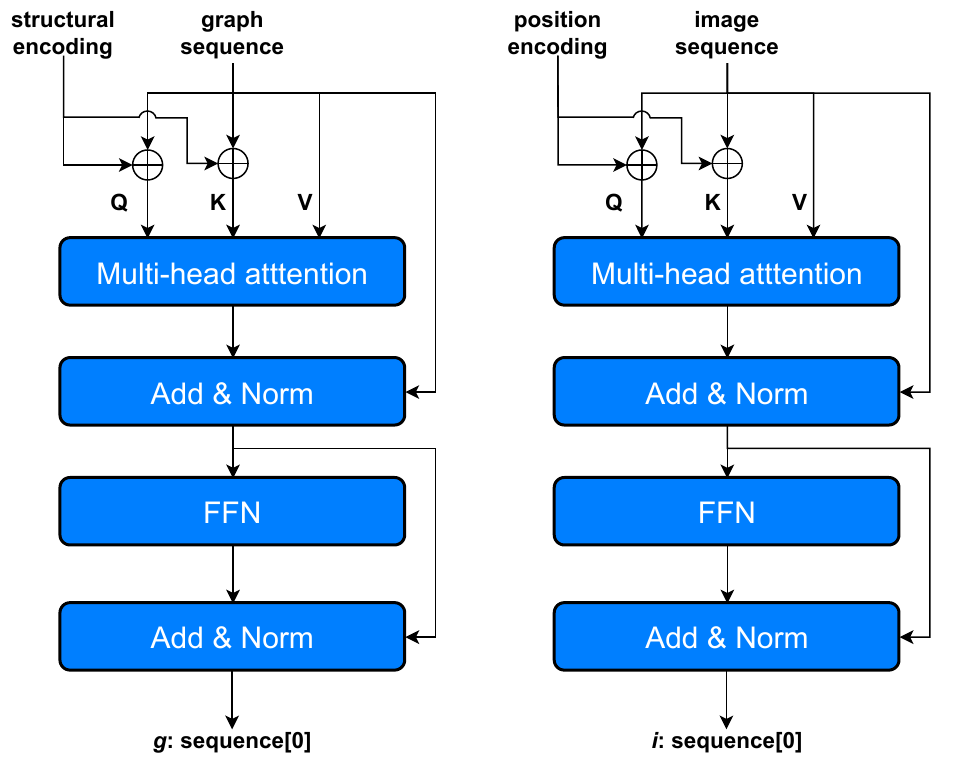}
\caption{Architecture of the graph Transformer and the image Transformer. 
The learnable graph embedding in the graph sequence from the last encoder layer is adopted as the final scene graph representation. This is also for images.
}
\label{fig:encoder}
\end{figure}

For both the graph Transformer and the image Transformer, we adopt the architecture of the vanilla Transformer encoder \cite{vaswani2017attention}, which is shown in~Fig.~\ref{fig:encoder}.
Each Transformer layer consists of a multi-head attention module, a feed-forward network, and two Layer normalization.
The graph embedding $\bm{g}$ and image embedding $\bm{i}$ from the last encoder layers are used as the final scene graph representation and image representation.

\subsection{Objective Function}
To align the scene graph and the corresponding image in the shared latent representation space, we adopt the contrastive loss as the optimization objective. 
Given a batch of $B$ scene graph-image pairs, the loss function is formulated as:

\begin{equation}
\centering
\begin{aligned}
L=-\frac{1}{B} \sum_{i=1}^B (\log\frac{\exp( \textrm{sim}(\bm{g}_i, \bm{i}_i))}{\sum_{j=1}^N \exp( \textrm{sim}(\bm{g}_i, \bm{i}_j))} \\
+ \log\frac{\exp( \textrm{sim}(\bm{g}_i, \bm{i}_i))}{\sum_{j=1}^N \exp( \textrm{sim}(\bm{g}_j, \bm{i}_i))}),
\end{aligned}
\label{eq:loss}
\end{equation}
where $\textrm{sim}(\bm{g}_i, \bm{i}_j)$ indicates the cosine similarity between the $i$-th scene graph representation $\bm{g}_i$ and the $j$-th image representation $\bm{i}_j$ in a batch. 
The first term in Eq.~\ref{eq:loss} aims to increase the similarity between the scene graph and its corresponding image compared to other images while the second term increases the similarity between the image and its scene graph compared to other scene graphs.

\section{Metrics for Scene Graph Generation}
\label{sec:metrics}

We review the common evaluation metrics for scene graph generation and their limitations. We further introduce R-Precision to evaluate the quality of generated scene graphs.

\subsection{Recall$@K$ and mean Recall$@K$}
Recall$@K$ is first introduced for visual relationship detection~\cite{lu2016visual} and is further widely used to evaluate SGG methods.
It calculates the fraction of ground truth triplets that appear in the top $K$ confident triplet predictions.
Due to the long tail issue in the scene graph datasets (\textit{e.g.}, Visual Genome~\cite{krishna2017visual}), mean Recall$@K$ is proposed to evaluate the performance of SGG models for low-frequency predicates~\cite{tang2019learning}.
It computes Recall$@K$ on each predicate category independently and takes their average.
However, both Recall$@K$ and mean Recall$@K$ have limitations since they strictly compare the predicted triplet set with the ground truth triplet set. 
Therefore, these metrics are not synonym-compatible and are sensitive to noise and bias in the dataset annotation.
Furthermore, they cannot directly demonstrate the alignment between the entire scene graph and the image.

\subsection{Mean Average Precision} 
mAP$_{rel}$ indicates the mean average precision of relationships where the predicted subject and object boxes have an IoU of at least 0.5 with ground truth boxes, while mAP$_{phr}$ is designed for the enclosing relationship boxes.
They are used to evaluate visual relationship detection in Open Images Challenge~\cite{kuznetsova2020open}.
Furthermore, wmAP$_{rel}$ and wmAP$_{phr}$ are proposed for more class-balanced evaluation~\cite{zhang2019graphical}.
However, such evaluation metrics are still at triplet-level and therefore have similar limitations as Recall$@K$ and mean Recall$@K$.

\subsection{R-Precision}

We observe the quality of a scene graph's description of an image is directly reflected in the degree of alignment within the shared latent space. 
To this end, we propose a novel evaluation metric R-Precision based on SPAN for scene graph generation evaluation. 
R-Precision measures the retrieval accuracy when retrieving the matching image from $K$ image candidates using the generated scene graph as a query.
We use the scene graph and image representations provided by SPAN to compute the scene graph-image similarity score for retrieval.
A larger value of $K$ indicates a more challenging retrieval task, demanding higher quality of the query scene graph.
Notably, our new metric is capable of evaluating both location-free and location-bound scene graphs.
In contrast to triplet-oriented metrics, R-Precision based on our framework compares the semantic context and visual appearance. Therefore, it is more robust to the perturbations from individual triplets.

\section{Experiments}
 \label{sec:exp}

\subsection{Datasets}
The experiments are conducted on two benchmark datasets for scene graph generation, i.e., Visual Genome \cite{krishna2017visual} and Open Images V6 \cite{kuznetsova2020open}. 
We establish new benchmarks with our graph-oriented metric for both datasets.

\noindent
\subsubsection{Visual Genome}
We follow the data split from \cite{xu2017scene}, which is adopted by most scene graph generation methods.
There are $108k$ images in the dataset with 150 entity categories and 50 predicate categories.
In the task of scene graph generation, $70\%$ of the images are used for training while the remaining $30\%$ are used for evaluation.

\subsubsection{Open Images V6}
Recently, an increasing number of scene graph generation methods have been validated on Open Images.
The large-scale Open Images V6 has 126$k$ training images and $5.3k$ test images with 300 entity
categories and 30 predicate categories.

\subsection{Technical Details}

\subsubsection{Node and Edge Representation}
We construct 150 entity embeddings as well as 50 predicate embeddings for Visual Genome~\cite{krishna2017visual}, and 300 entity embeddings as well as 30 predicate embeddings for Open Images V6 \cite{kuznetsova2020open}. 
They are learnable embeddings according to the entity or predicate categories.
For a location-free scene graph, the node representations are the corresponding entity embeddings and the edge representations are the corresponding predicate embeddings.
For a location-bounded scene graph, the node representations are the corresponding entity embeddings element-wise added with the box embeddings, which are inferred from the bounding box coordinates.

\subsubsection{Box Embedding}
\begin{figure}[htp!]
\centering
\includegraphics[width=1\linewidth]{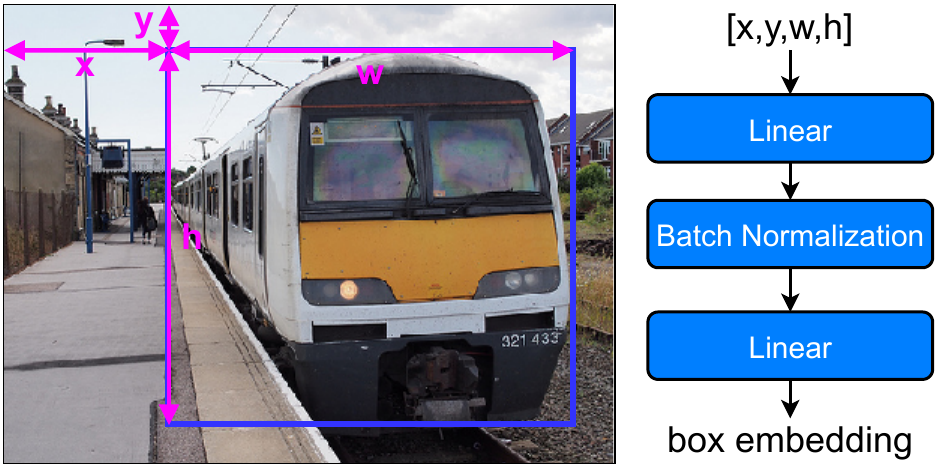}
\caption{We convert the bounding box coordinates $[x,y,w,h]$ into the box embedding with linear layers with batch normalization, where $x,y$ represent the coordinates of the upper left corner of the box, while $w, h$ denote the width and height, respectively.
}
\label{fig:box_embed}
\end{figure}

The box embedding has a dimension of 512, which is the same as the node representation.
This allows the two to be element-wise added together.
For a location-bounded scene graph, the given bounding box coordinates $[x,y,w,h]$ of a node are transformed into a box embedding by linear transformation layers with normalization, as depicted in~Fig.~\ref{fig:box_embed}. 
The box embedding is then injected into the corresponding node representation as location information.

\subsubsection{Graph Serialization}
\begin{figure*}[htp!]
\centering
\includegraphics[width=0.85\linewidth]{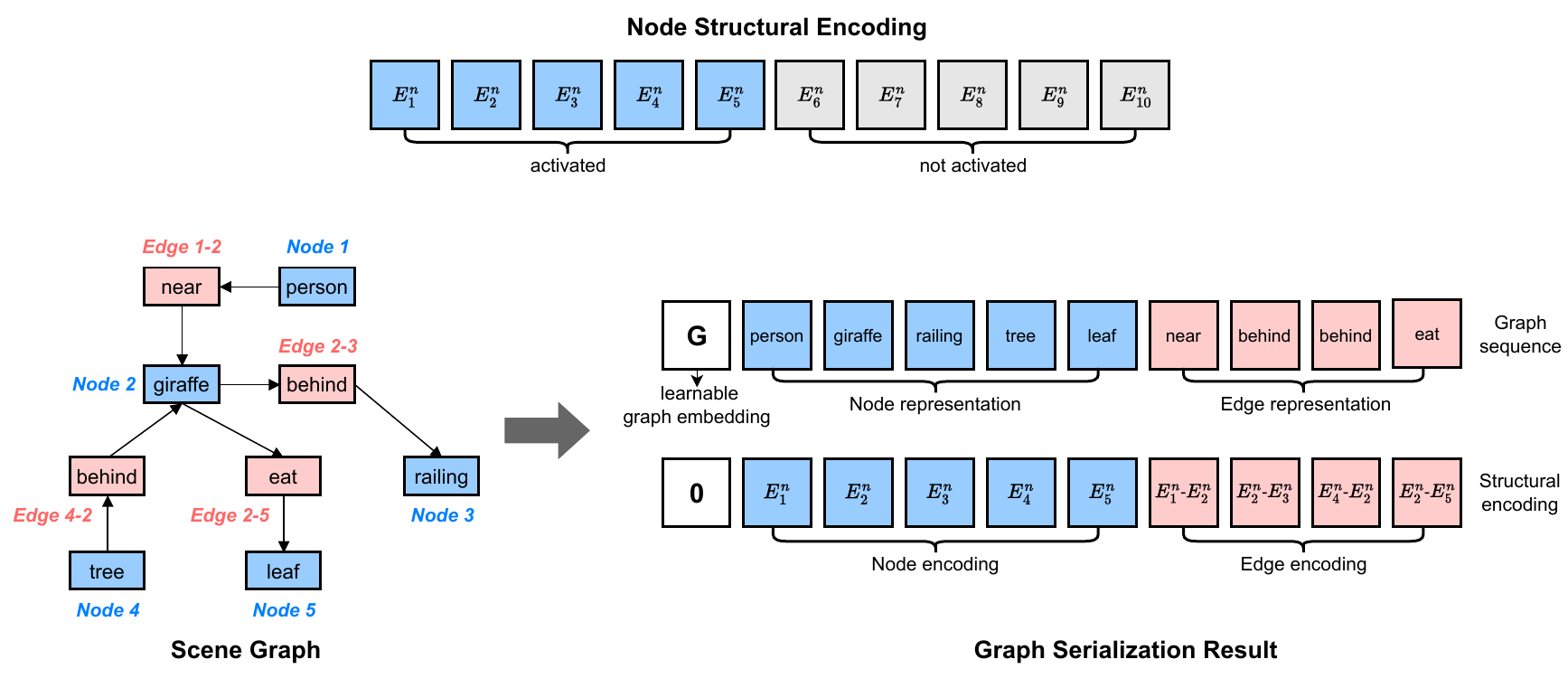}
\caption{An example of graph serialization. The node structure encoding is randomly assigned to the node representation, while the edge encoding is calculated based on the start node encoding and the end node encoding.
}
\label{fig:graph_serial}
\end{figure*}

To enable the Transformer to handle a scene graph, we propose a graph serialization technique, where a scene graph is serialized to a sequence of node representations and edge representations.
In order to retain the structural information, we introduce structural encoding to represent the graphical structure, which is further used in the graph Transformer.
An example of graph serialization is illustrated in~Fig.~\ref{fig:graph_serial}.
We first construct 10 learnable node structural encodings to distinguish different nodes in the scene graph.
If a scene graph contains more than 10 entity nodes, we randomly sample 10 from the nodes and clip the others. 
If the number of nodes is less than 10, the redundant node encodings are not activated.
The node encodings are randomly assigned to the nodes.
In the given example, we view \texttt{[person]} as the first node and assign $\bm{E}^n_1$ to it.
\texttt{[giraffe]} is viewed as the second node, and so on.
The edge encoding indicates from which node the corresponding edge is directed to which node.
For example, the edge encoding for \texttt{[behind]} from the second node \texttt{[giraffe]} to the third node \texttt{[railing]} is $\bm{E}^n_2-\bm{E}^n_3$.
For another \texttt{[behind]} in the scene graph, the corresponding encoding is $\bm{E}^n_4-\bm{E}^n_2$, since it starts from the fourth node \texttt{[tree]} to the second node \texttt{[giraffe]}.
To capture the context and learn the graph representation, we insert a learnable graph embedding in the graph sequence but do not preserve an encoding for it.

\subsubsection{CLIP baseline for image retrieval}
\begin{figure}[tp]
\centering
\includegraphics[width=0.95\linewidth]{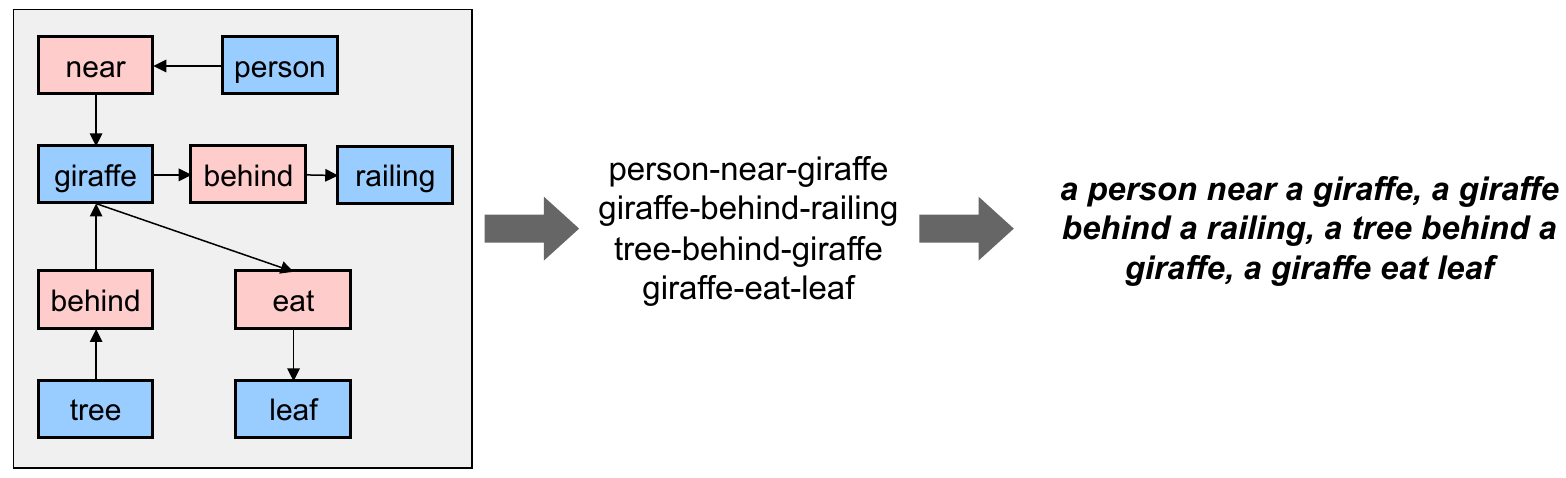}
\caption{To convert the location-free scene graph into a sentence that can be used as a query for image retrieval, we combine the triplets within the scene graph and incorporate indefinite articles to specify the entities.
}
\label{fig:clip}
\end{figure}

For the task of image retrieval, we employ CLIP (ViT-B-32) \cite{radford2021learning} to establish a baseline.
Since CLIP only supports text as input, we concatenate the triplets in the location-free scene graph into a string format (see~Fig.~\ref{fig:clip}).
The converted text and images are respectively encoded by the CLIP text encoder and image encoder.
The similarity between query text and image candidates is computed for image retrieval.
Compared to CLIP, our framework SPAN performs better, which uses location-free scene graphs as queries.

\subsubsection{Further Implementation Details}
We adopt the same setting for training on both the Visual Genome and Open Images V6 datasets.
We train our framework using an AdamW optimizer \cite{loshchilov2017decoupled} with a learning rate of $1\times e^{-4}$ on 8 GPUs with a batch size of 32 per GPU. 
The model dimension $d$ for both the graph Transformer and image Transformer is set to 512, allowing for the calculation of similarity between scene graphs and images.
If not  specified otherwise, the default number of layers for the graph Transformer and the image Transformer is 6. 
The length of the learnable node structural encodings is restricted to 10. 
If a scene graph contains more than 10 entity nodes, we randomly sample 10 nodes and clip the others. 
The input image is scaled, with the longest side being 512, and is zero-padded into a square.
For both datasets, only the training set is used in the contrastive training.

\begin{table*}[t!]
\centering
\begin{adjustbox}{width=1\textwidth}
\begin{tabular}{c|c|cccc|ccc|ccc}
 \hline \hline
 \multicolumn{2}{c|}{\multirow{2}*{Method}}& \multicolumn{4}{c|}{SGDET} & \multicolumn{3}{c|}{R-Precision (LF Graph)} &\multicolumn{3}{c}{R-Precision (LB Graph)} \\
  \multicolumn{2}{c|}{} & R@20 &R@50 &mR@20 &mR@50 & K=10 & K=50 & K=100 & K=10 & K=50 & K=100 \\
  \hline 
\multicolumn{2}{c|}{Ground Truth}  & - & - & - &- & 86.7 &67.9 &58.0 &92.1 &78.9 &71.0 \\
\hline 
 \multirow{9}*{\makecell[c]{two-\\stage}}
& MOTIFS \cite{zellers2018neural}& 21.4 & 27.2 & \underline{4.1} & \underline{5.7} &84.5 &65.1 &54.9 &90.2 &74.5 &69.1  \\
& RelDN \cite{zhang2019graphical}&21.1 &28.3 & \underline{4.9} & \underline{6.4} &88.9 &72.5 &62.7 &93.5 &83.1 &75.0\\
& NODIS \cite{cong2020nodis}& 21.6 & 27.7 & \underline{5.1} & \underline{6.0}  &85.0 &66.9 &58.1 &90.5 &75.6 &69.8 \\
& MOTIFS-TDE$^\dagger$ \cite{tang2020unbiased} & 12.4 & 16.9 & 5.8 & 8.2 & 94.5 &82.9 &74.7 &\textbf{97.8} &91.8 & 87.1 \\
& VCTree-EBM$^\dagger$ \cite{suhail2021energy} & 24.2 &31.4 & 5.7 & 7.7 &92.9 &78.0 &68.9 &96.5 &87.7 &81.7 \\
& BGNN \cite{li2021bipartite}& \underline{23.3} & 31.0 & \underline{7.5} & 10.7  &93.3 &79.4 &70.9 &97.1 &89.7 &84.3\\
& SHA-GCL$^\dagger$ \cite{dong2022stacked}& \underline{11.9} & 14.9 & \textbf{14.2} & \textbf{17.9}  &94.7 &82.7 &72.9 &97.7 &91.7 &86.5 \\
& FGPL$^\dagger$ \cite{lyu2022fine}& \underline{21.0} & \underline{25.7} & 13.2 & 17.4  & \textbf{95.0} & \textbf{83.6} & \textbf{75.6} & \textbf{97.8} & \textbf{92.1} &\textbf{87.6} \\
& MOTIFS-NICE$^\dagger$ \cite{li2022devil}& \underline{20.4} & 27.8 & \underline{7.4} & 12.2  &94.3 &82.3 &71.2 &97.1 &90.0 &84.6\\
\hline
 \multirow{4}*{\makecell[c]{one-\\stage}}
&FCSGG \cite{liu2021fully}&  16.1 & 21.3 & 2.7 & 3.6 &79.0 &59.2 &49.8 &87.4 &70.1 &66.7 \\
&SSR-CNN \cite{teng2022structured}& \textbf{25.8} & \textbf{32.7} & 6.1 & 8.4 &91.6 &74.4 & 64.0 &95.9 &86.5 &79.6 \\
&SGTR \cite{li2022sgtr}& \underline{19.5} & 24.6 & \underline{7.5} & 12.0 &94.5 & 83.1 & 74.9 &97.3 &90.8 &85.7  \\
&RelTR \cite{10105507}& 21.2 & 27.5 & 6.8 & 10.8 &93.1 &80.1 &71.1 &97.0 &90.1 &84.8 \\
\hline \hline
\end{tabular}
\end{adjustbox}
\caption{Benchmarking different scene graph generation models on the Visual Genome dataset \cite{krishna2017visual}.
We re-evaluate 9 two-stage methods and 4 one-stage methods using R-Precision (K=10/50/100) for location-free scene graphs (LF Graph) and location-bounded scene graphs (LB Graph). 
The unbiased SGG methods are labeled with $\dagger$. 
The highest score in each column is in \textbf{bold}. The underlined \underline{numbers} indicate that the scores are not available in the original papers, but are re-evaluated in this work.}
\label{tab:vg_benchmark}
\end{table*}

\subsection{Scene Graph Generation Benchmarks}
Considering that the triplet-oriented metrics cannot directly reflect the consistency between generated scene graphs and images,
we re-evaluate the prior scene graph generation models with the graph-oriented metric R-Precision (K$=10/50/100$) based on the similarity computed by SPAN.
We create two benchmarks, respectively for the Visual Genome and Open Images V6 datasets, and perform a systematic analysis.
Different from triplet-oriented metrics, which require bounding box information to calculate IoU, the new metric is also applicable to location-free scene graphs and provides a reference for semantic downstream tasks.
Despite the large amount approaches for scene graph generation,  we only evaluate a selection of representative models for which official code and trained weights are available due to limited computation resources.

\subsubsection{Visual Genome Benchmark}
We benchmark 9 two-stage SGG methods~\cite{zellers2018neural, zhang2019graphical, cong2020nodis, tang2020unbiased, suhail2021energy, li2021bipartite, lyu2022fine, li2022devil, dong2022stacked} and 4 one-stage SGG methods~\cite{liu2021fully, li2022sgtr, 10105507, teng2022structured} for Visual Genome. 
Some of them are specific for unbiased scene graph generation~\cite{tang2020unbiased,suhail2021energy,lyu2022fine,dong2022stacked,li2022devil}.
The evaluation scores are given in Tab.~\ref{tab:vg_benchmark}.
For both location-free scene graphs and location-bounded scene graphs, we are surprised to find that R-Precision of most methods is \textbf{higher} than R-Precision using ground truth scene graphs (first row in Tab. \ref{tab:vg_benchmark}).
For location-free scene graphs, the R-Precision (K=100) of FGPL \cite{lyu2022fine} is $17.6$ higher than the ground truth, and for location-bounded scene graphs, the R-Precision (K=100) of FGPL is $16.6$ higher than the ground truth.
This demonstrates that the scene graphs generated by the existing models have higher similarity to the corresponding images than manually labeled scene graphs and perform better in the task of image retrieval.
We conjecture that the main reasons are as follows: (1) The manual annotations of Visual Genome are noisy. Some ground truth scene graphs consist of only one or two relationships, while generated scene graphs include more triplets. 
(2) The problem of bias in ground truth scene graphs has been of much concern.  Several models have demonstrated the ability to generate unbiased scene graphs even when trained on biased data. 
This increases the diversity of information in the generated scene graphs.

\begin{figure}[tp!]
\centering
\includegraphics[width=1\linewidth]{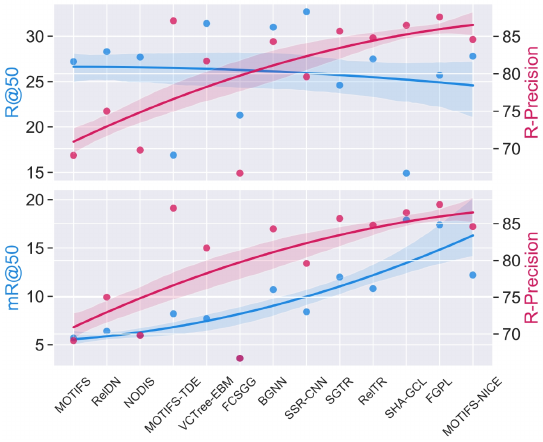}
\caption{R$@50$, mR$@50$ and R-precision (K=100) scores of different scene graph generation models on the Visual Genome dataset for location-bounded scene graphs. The models on the X-axis are sorted by the time they were released on the arXiv.}
\label{fig:rcurve}
\end{figure}

Moreover, we find that recent models have been unable to achieve breakthroughs in Recall$@K$ (R$@K$) (see~Fig.~\ref{fig:rcurve}). 
Therefore more unbiased SGG works have emerged, which focus on mean Recall$@K$ (mR$@K$). 
However, some of them have high mR$@K$ but low R$@K$.  
The different performance of the same model on these two recall metrics makes the model performance unreliable.
Our metric R-Precision reflects the performance of the SGG models in a different way by evaluating the alignment between the generated scene graphs and the images. 
This complements and strengthens the existing SGG evaluation metrics.

\begin{table*}[t!]
\centering
\begin{adjustbox}{width=1\textwidth}
\begin{tabular}{c|cccc|ccc|ccc}
 \hline \hline
\multirow{2}*{Method} & \multirow{2}*{R@50} & \multirow{2}*{wmAP$_{rel}$} & \multirow{2}*{wmAP$_{phr}$} & \multirow{2}*{score$_{wtd}$} & \multicolumn{3}{c|}{R-Precision (LF Graph)} &\multicolumn{3}{c}{R-Precision (LB Graph)} \\
& & & & & K=10 & K=50 & K=100 & K=10 & K=50 & K=100 \\
\hline 
Ground Truth & - & - & - &- & 89.2 &67.5 &55.4 &94.2 &82.0 &73.4   \\
\hline 
BGNN \cite{li2021bipartite}& 75.0 & 33.5 & 34.2 & 41.7  &72.3 &43.1 &32.6 &91.5 & 75.6 & 66.8  \\
SGTR \cite{li2022sgtr}& 59.9 & 37.0 & 38.7 & 42.3 &\textbf{82.6} & \textbf{57.6} &\textbf{46.3} & 94.9 &83.6 &76.0  \\
RelTR \cite{10105507}& 71.7 & 34.2 & 37.5 & 43.0 &79.9 &55.0 &42.1 &93.9 & 82.1 &75.2 \\
SSR-CNN \cite{teng2022structured}& \textbf{76.7} & \textbf{41.5} & \textbf{43.6} & \textbf{49.4} & 77.7 &53.3 &40.9& \textbf{95.9} & \textbf{85.6} & \textbf{78.7}  \\
\hline \hline
\end{tabular}
\end{adjustbox}
\caption{Benchmarking different scene graph generation models on the Open Images V6 dataset \cite{kuznetsova2020open}.
We re-evaluate 4 scene graph generation methods using R-Precision (K=10/50/100) for location-free scene graphs (LF Graph) and location-bounded scene graphs (LB Graph). The highest score in each column is in \textbf{bold}.}
\label{tab:oi_benchmark}
\end{table*}

\begin{figure*}[ht!]
\centering
\includegraphics[width=1\linewidth]{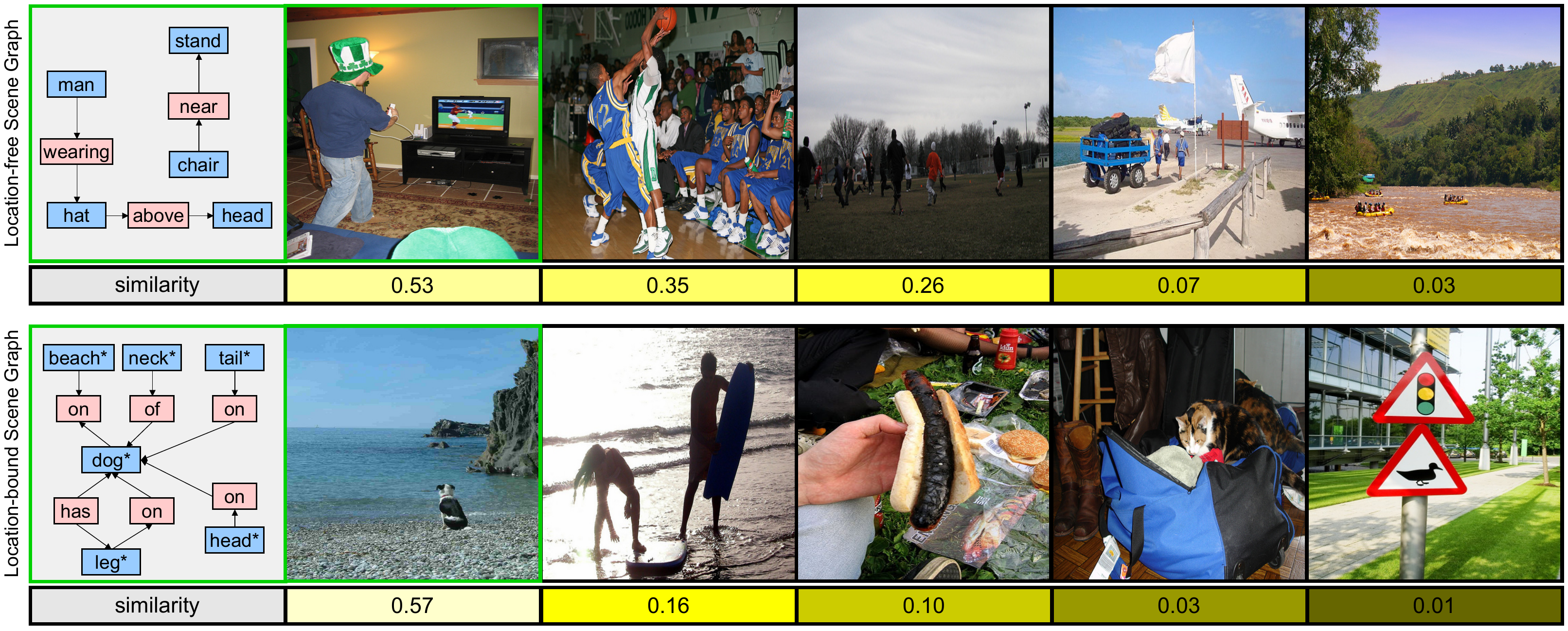}
\caption{Qualitative results of similarity between scene graphs and images on the Visual Genome dataset.
Our framework can infer reasonable similarities between images and location-free as well as location-bounded scene graphs for image retrieval.
The matched scene graphs and images are highlighted in green.
\texttt{entity$^*$} indicates the bounding box is given for the location-bounded scene graph. 
}
\label{fig:retrieval3}
\end{figure*}

\subsubsection{Open Images V6 Benchmark}
We benchmark 4 SGG methods \cite{li2021bipartite,li2022sgtr,10105507,teng2022structured} for the Open Images V6 dataset.
The evaluation scores are shown in Tab.~\ref{tab:oi_benchmark}, where score$_{wtd}$ indicates the weighted $\text{score}=0.2\times \text{R}@50+0.4\times \text{wmAP}_{rel}+0.4\times \text{wmAP}_{phr}$.
The R-Precision scores using ground truth scene graphs for retrieval on the Open Images dataset are similar to Visual Genome. 
The R-Precision (K=100) is $55.4$ for location-free scene graphs and $73.4$ for location-bounded scene graphs.
We observe that the SGG models do not achieve the same level of performance on Open Images as they do on Visual Genome, especially for location-free scene graphs.
The reason for this could be that Open Images V6 has a different data distribution (e.g. more entity categories) and the optimization of SSG model hyperparameters is usually based on the Visual Genome dataset.
For location-free scene graphs, SGTR \cite{li2022sgtr} has the best performance with  R-Precision (K=100) of $46.3$, which is $9.1$ lower than using ground truth scene graphs.
For location-bounded scene graphs, SSR-CNN \cite{teng2022structured} outperforms the other methods with R-Precision (K=100) of $78.7$, which is $5.3$ higher than using ground truth scene graphs.
Due to limited data, it is difficult to determine the correlation between R-Precision and wmAP$_{rel}$/wmAP$_{phr}$.
However, for location-bounded scene graphs, the higher the weighted mAP of the model, the higher its R-Precision score.

\begin{table}[t!]
\centering
\begin{adjustbox}{width=0.49\textwidth}
\begin{tabular}{c|ccc|ccc}
 \hline \hline
\multirow{2}*{Method} & \multicolumn{3}{c|}{R-Precision (LF Graph)} &\multicolumn{3}{c}{R-Precision (LB Graph)}\\
 & K=10 & K=50 & K=100 & K=10 & K=50 & K=100 \\
\hline 
CLIP \cite{radford2021learning} &79.8 &57.8 &47.2 &- &- &- \\
\hline 
SPAN-Node &78.9 &56.6 & 46.4 &84.2 &70.8 & 64.6 \\
SPAN-Edge &12.1 & 3.4 &1.8 & - &- &- \\
SPAN &86.7 &67.9 &58.0 &92.1 &78.9 &71.0 \\
\hline \hline
\end{tabular}
\end{adjustbox}
\caption{Quantitative results for image retrieval task using R-Precision as the evaluation metric.
Compared to CLIP, SPAN has higher R-Precision by using scene graphs as the query.
SPAN-Node and SPAN-Edge indicate that only node embeddings or edge embeddings are given to compute the similarity.
}
\label{tab:quantitative_result}
\end{table}

\subsection{Similarity between Scene Graphs and Images}
\label{sec: similarity}
\begin{figure*}[htp!]
\centering
\includegraphics[width=0.7\linewidth]{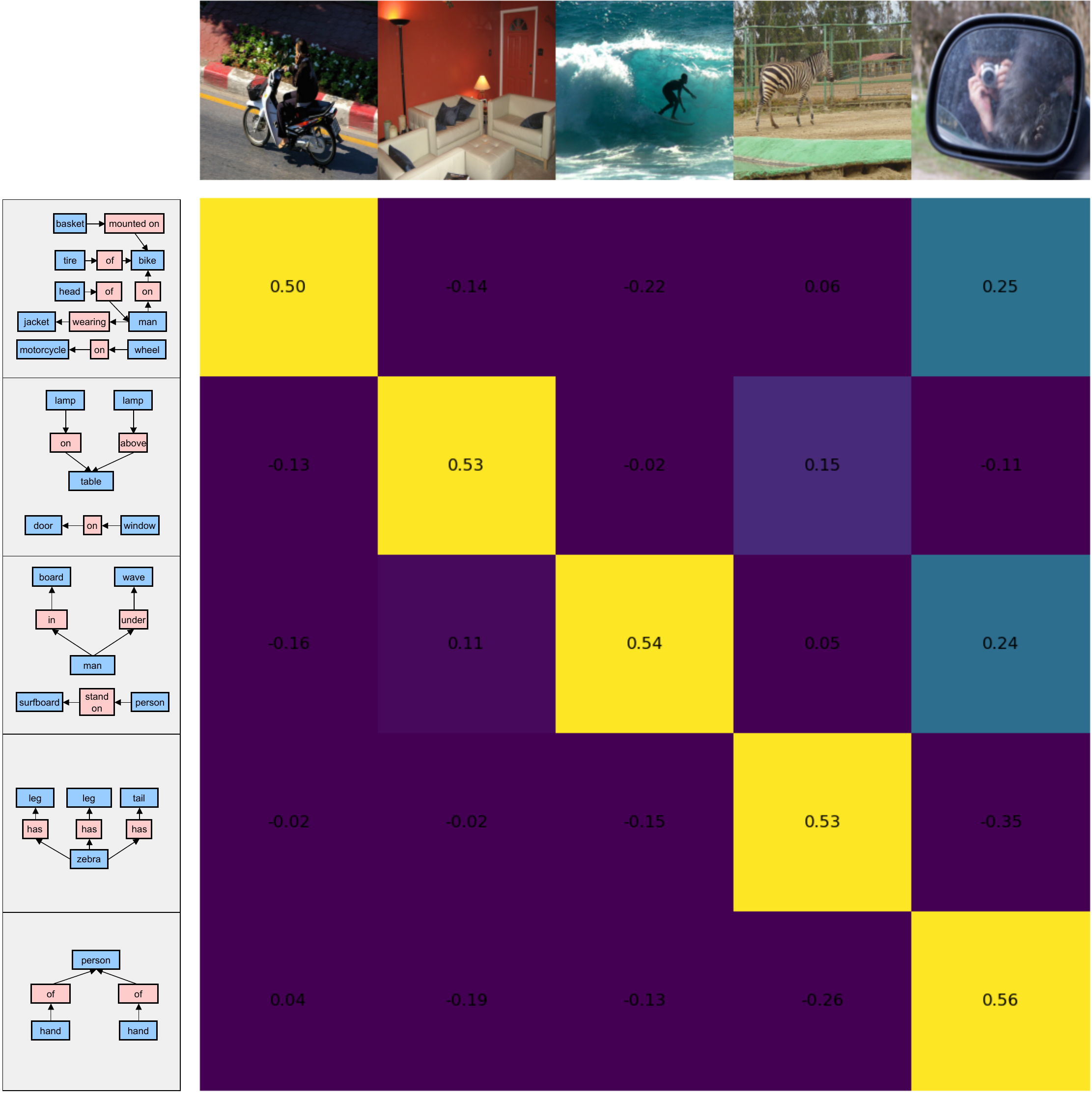}
\caption{Qualitative results of similarity between location-free scene graphs and images for the Visual Genome dataset.
For better visualization, the images are resized.
}
\label{fig:vg_qual1}
\end{figure*}

\begin{figure*}[htp!]
\centering
\includegraphics[width=0.7\linewidth]{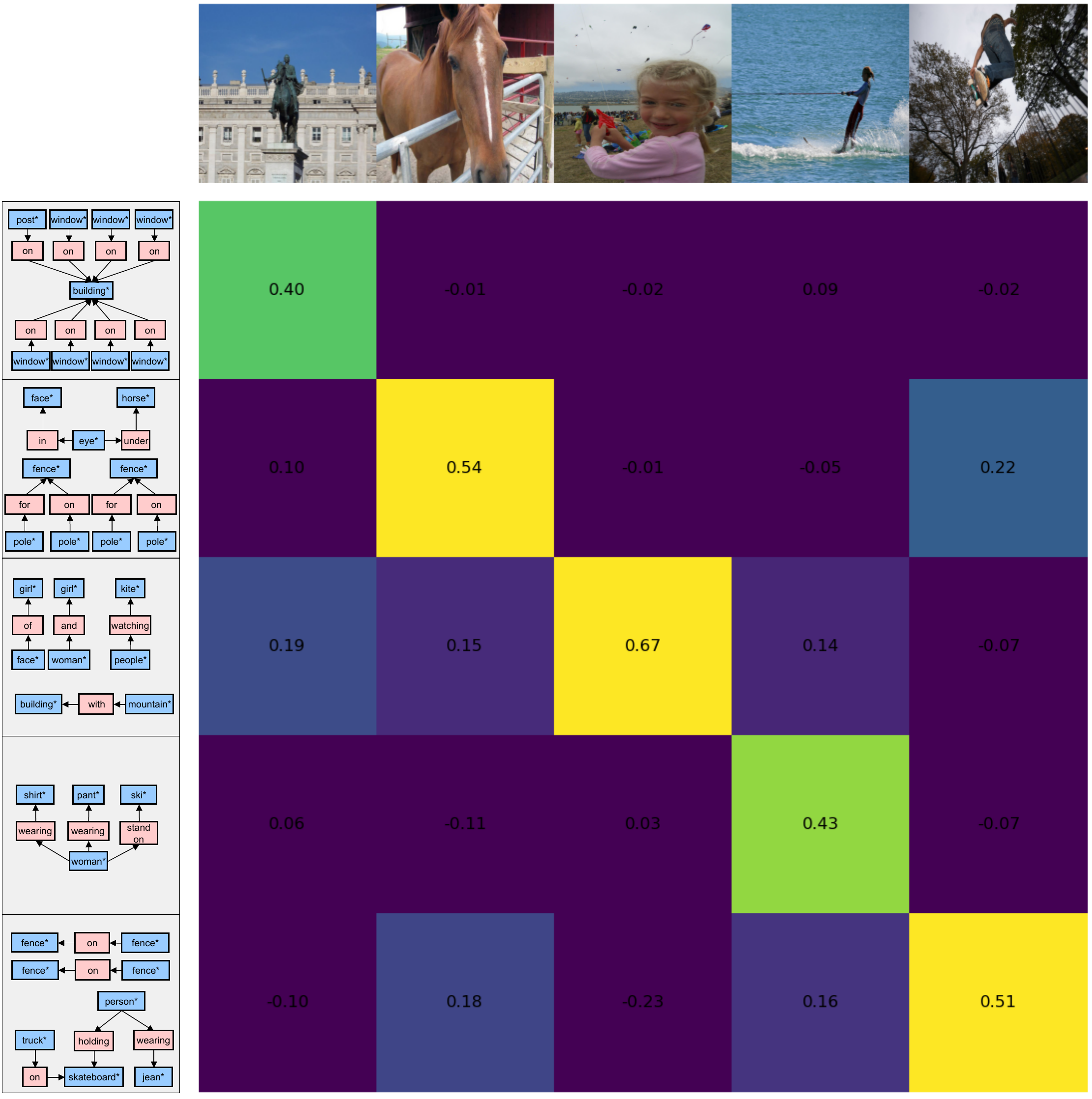}
\caption{Qualitative results of similarity between location-bounded scene graphs and images for the Visual Genome dataset.
For better visualization, the images are resized.
\textit{entity}* indicates the bounding box is given in the location-bounded scene graph.
}
\label{fig:vg_qual2}
\end{figure*}

\begin{figure*}[htp!]
\centering
\includegraphics[width=0.7\linewidth]{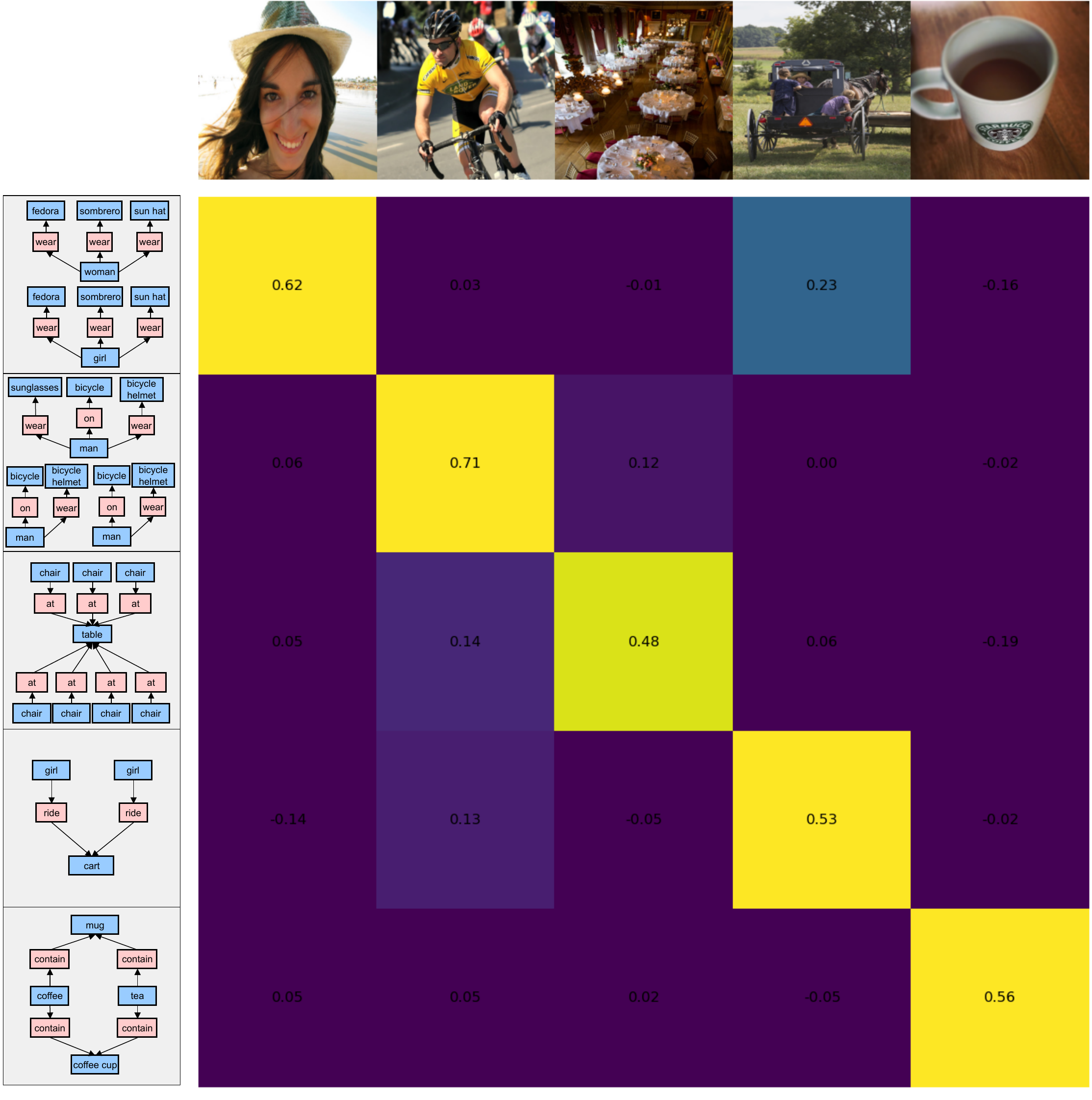}
\caption{Qualitative results of similarity between location-free scene graphs and images for the Open Images V6 dataset.
For better visualization, the images are resized.
}
\label{fig:oi_qual1}
\end{figure*}

\begin{figure*}[htp!]
\centering
\includegraphics[width=0.7\linewidth]{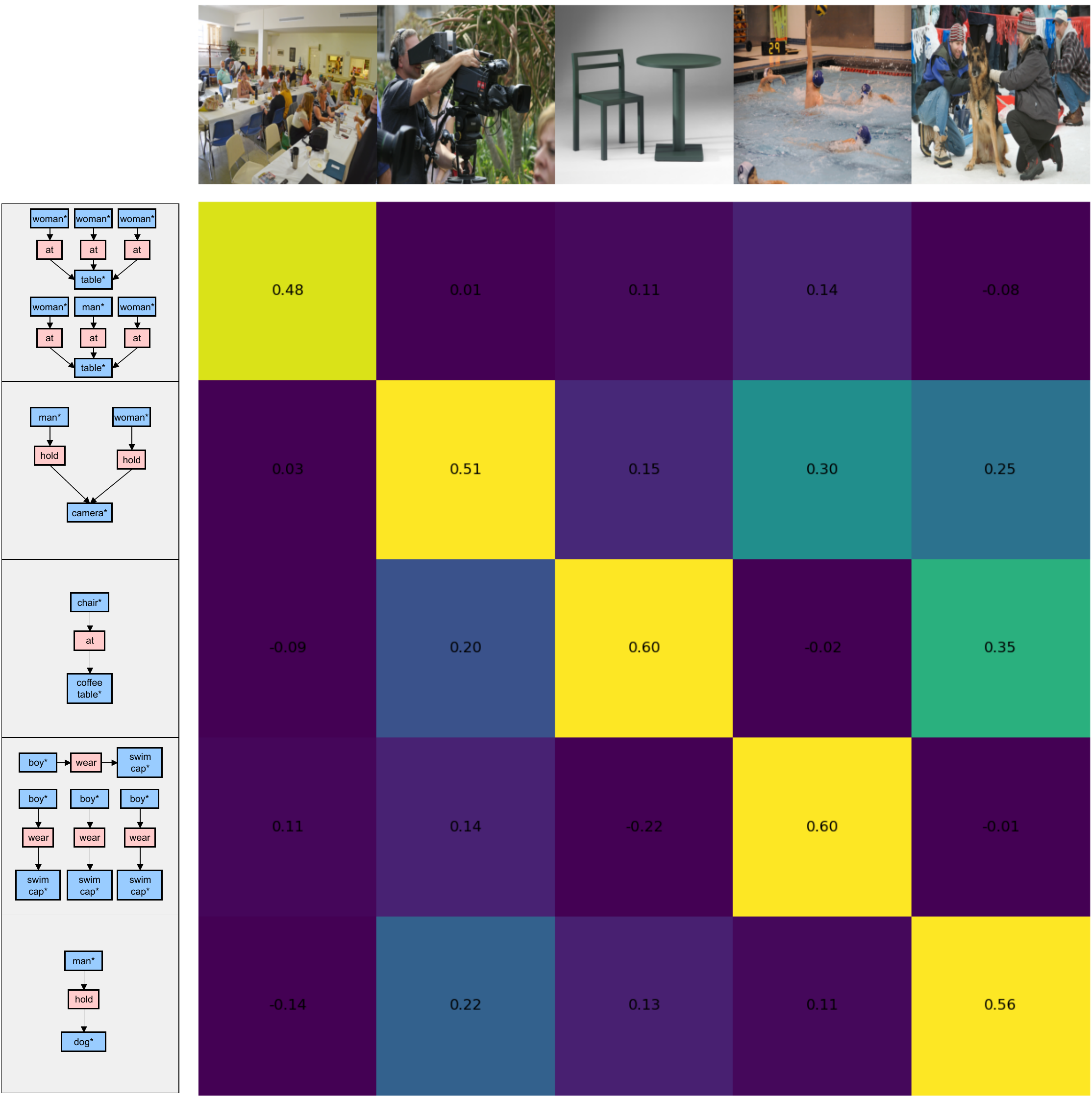}
\caption{Qualitative results of similarity between location-bounded scene graphs and images for the Open Images V6 dataset.
For better visualization, the images are resized.
\textit{entity}* indicates the bounding box is given in the location-bounded scene graph.
}
\label{fig:oi_qual2}
\end{figure*}

We compute the cosine similarity between scene graphs and images. 
For the Visual Genome dataset, the maximum similarity between any location-free/location-bounded scene graph and image in the test set is 0.79/0.78 and the minimum is -0.55/-0.59, while the average similarity between matched ground truth scene graphs and images in the test set is 0.50/0.55.
The qualitative results for similarity calculation are shown in~Fig.~\ref{fig:vg_qual1} (using location-free scene graphs) and in~Fig.~\ref{fig:vg_qual2} (using location-bounded scene graphs).
For the Open Images V6 dataset, the maximum similarity between any location-free/location-bounded scene graph and image in the test set is 0.83/0.76 and the minimum is -0.58/-0.53, while the average similarity between matched ground truth scene graphs and images in the test set is  0.56/0.56.
The qualitative results are shown in Fig.~\ref{fig:oi_qual1} (using location-free scene graphs) and in~Fig.~\ref{fig:oi_qual2} (using location-bounded scene graphs).



\subsection{Image Retrieval with Similarity Scores}

To evaluate the effectiveness of our framework, we conduct image retrieval experiments using the similarity scores between scene graphs and images provided by SPAN on the Visual Genome dataset.

We first employ CLIP (ViT-B-32) \cite{radford2021learning} to establish a baseline. 
Considering that CLIP only encodes text, we concatenate the triplets in the scene graph into a string format and use the text as the query for image retrieval. More details are provided in the supplementary material. 
Since CLIP does not incorporate bounding box information, we only test location-free scene graphs using this method.
Furthermore, we conduct experiments to investigate the contribution of node and edge information in scene graphs to the image retrieval task. 
We use only node (SPAN-Node) or edge (SPAN-Edge) information as the query for image retrieval, and for both methods, the input graph structure is artificially disrupted. 
During training and evaluation, only node embeddings or edge embeddings are given.

The quantitative results are shown in Tab.~\ref{tab:quantitative_result}. 
Compared to CLIP, SPAN has significantly better R-Precision.
In addition, SPAN can retrieve images more accurately with the additional constraint of entity bounding boxes.
Specifically, when using location-bounded scene graphs as queries, R-Precision (K=100) is $13.0$ higher than using location-free scene graphs.
Comparing SPAN-Node and SPAN-Edge, the nodes in the scene graph, \textit{i.e.}, the entities in the scene, play a dominant role in image retrieval. 
This observation underscores the significance of entities in defining the core information within a visual scene.
Adding the edge information helps understand the visual scene structurally and improves image retrieval performance.



Fig.~\ref{fig:retrieval3} shows the qualitative results of image retrieval using the similarity scores between scene graphs and images. 
The matched pairs are highlighted in green and the similarity values are not scaled.
We use location-free scene graphs, and location-bounded scene graphs as queries to retrieve the matching image from 100 image candidates on the Visual Genome dataset.
The query scene graph and the image candidates are encoded, and the similarity between the scene graph and images is computed.
The candidate with the highest similarity is viewed as the matched image.
Our framework, SPAN, is capable of inferring reasonable similarities between images and both location-free and location-bounded scene graphs, leading to successful image retrieval.  
Moreover, we note that the use of location-bounded scene graphs, which include the constraint of entity boxes, provides enhanced discrimination of visual scenes compared to employing location-free scene graphs.

\begin{figure*}[htp!]
\centering
\includegraphics[width=1\linewidth]{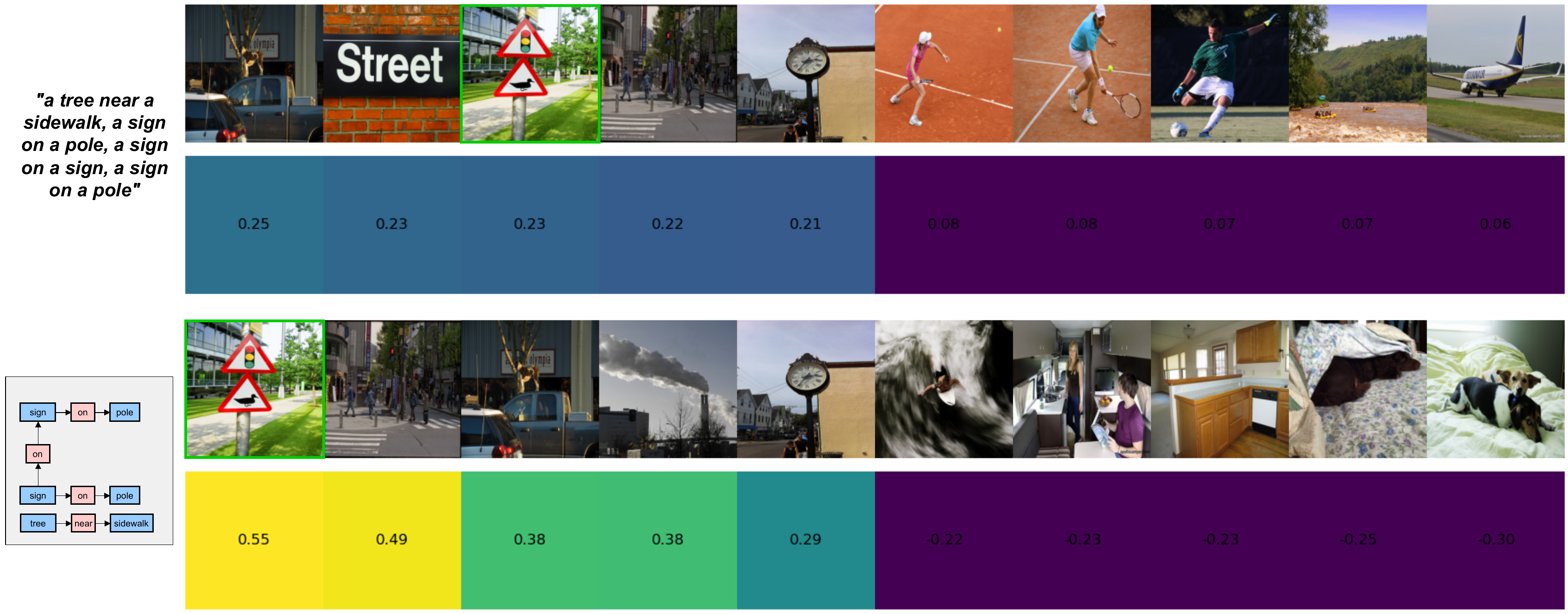}
\caption{Qualitative results for image retrieval using the text (first row) and location-free scene graph (second row) as queries. The five images with the highest similarity and the five images with the lowest similarity from 100 image candidates are shown.
The corresponding image to the query is highlighted in green.
CLIP cannot retrieve the correct image, while our framework assigns the highest similarity score to the matched image, indicating its effectiveness in image retrieval.
}
\label{fig:vg_r1}
\end{figure*}

\begin{figure*}[htp!]
\centering
\includegraphics[width=1\linewidth]{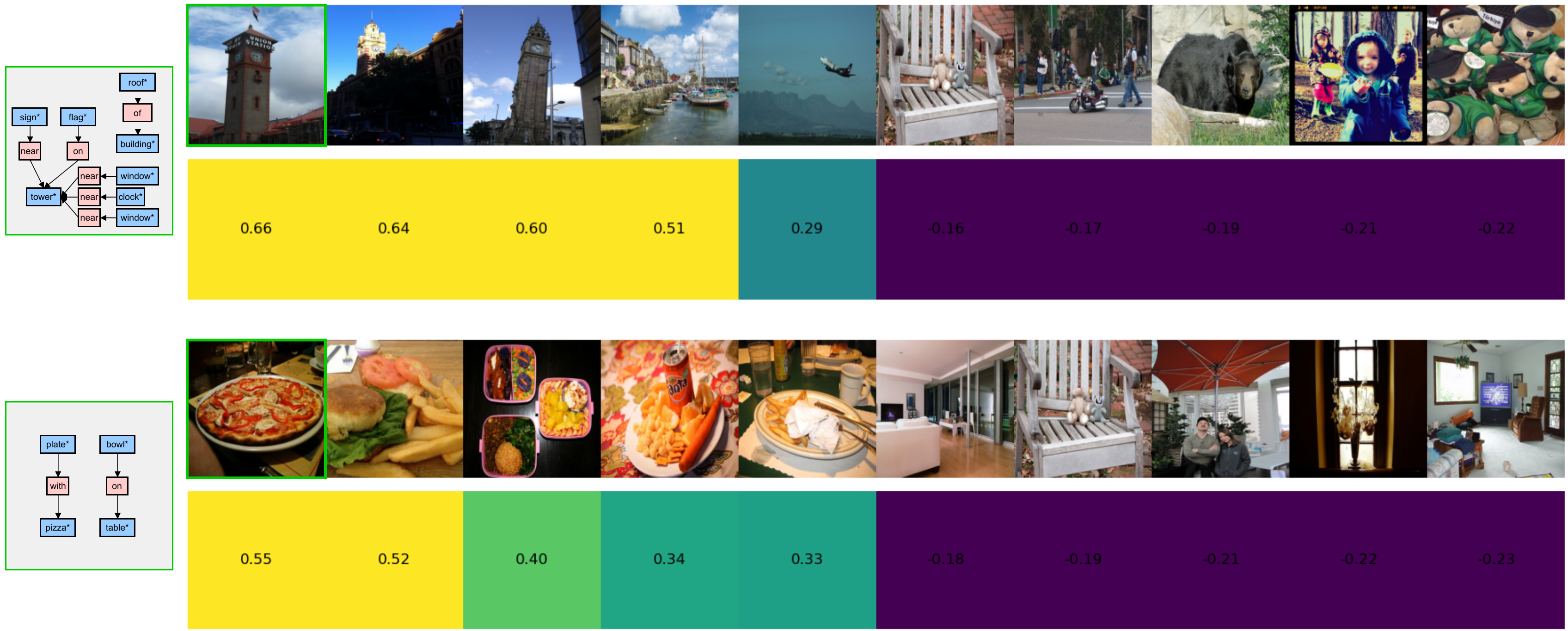}
\caption{
Qualitative results for image retrieval using location-bounded scene graphs as queries. The five images with the highest similarity and the five images with the lowest similarity from 100 image candidates are shown.
The images corresponding to the scene graphs in the ground
truth are highlighted in green. 
\textit{entity}* indicates the bounding box is given in the location-bounded scene graph.
}
\label{fig:vg_r2}
\end{figure*}

Further qualitative results using the text and location-free scene graph are presented in~Fig.~\ref{fig:vg_r1}, and the qualitative results using location-bounded scene graphs are presented in~Fig.~\ref{fig:vg_r2}. 
For comparison, we also use CLIP to retrieve the image that match the text.
We show the five images with the highest similarity and the five images with the lowest.
The image corresponding to the input query in the ground truth is highlighted in green.
Due to the large number of image candidates, some scenes similar to those described in the scene graph appear in the top five images.

\begin{table}[t!]
\centering
\begin{adjustbox}{width=0.48\textwidth}
\begin{tabular}{c|ccc|ccc|c}
 \hline \hline
\multirow{2}*{Method} & \multicolumn{3}{c|}{R-Precision (LF Graph)} &\multicolumn{3}{c|}{R-Precision (LB Graph)} & \multirow{2}*{\# param}\\
 & K=10 & K=50 & K=100 & K=10 & K=50 & K=100 & \\
\hline 
GCN &78.9 &57.9 &46.2 &84.5 & 70.9 &63.2 & 46.2M \\
\hline 
w/o SE  &80.1 & 58.1 & 47.8 &87.2 &71.7 &65.1  & \multirow{3}*{36.9M}\\
w/o NS  &83.4 &62.4 &52.8 &89.4 &75.0 &68.3  & \\
full &\textbf{86.7} &\textbf{67.9} &\textbf{58.0} &\textbf{92.1} &\textbf{78.9} & \textbf{71.0} &  \\
\hline \hline
\end{tabular}
\end{adjustbox}
\caption{Ablation study of the Transformer-based graph encoder. We compare the effectiveness of the graph Transformer and the baseline graph convolutional network (GCN). Moreover, we ablate structural encodings (SE) and node shuffle (NS) in graph serialization to analyze their impact.}
\label{tab:ablate_transformer}
\end{table}


\subsection{Ablation Studies}

\subsubsection{Number of Layers}

We conduct experiments on the Visual Genome dataset to verify the impact of different layer numbers on the performance of the graph Transformer and the image Transformer. 
The results are presented in~Tab.~\ref{tab:ablate_transformer}.
For location-free scene graphs, when both the graph Transformer and the image Transformer have 4 layers, the R-precision score (K$=100$) is 54.9. For location-bounded scene graphs, the score is 66.5. As we increase the number of layers in both Transformers, the model performance improves due to the increased number of parameters.
The best performance is achieved when both Transformers have 6 layers. However, increasing the number of layers to 8 leads to a drop in performance. 
We conjecture that this may be caused by overfitting.


\begin{table}[tp]
\centering
\begin{adjustbox}{width=0.49\textwidth}
\begin{tabular}{cc|cc|cc|c}
 \hline \hline
\multicolumn{2}{c|}{Layer Number} & \multicolumn{2}{c|}{R-Precision (LF Graph)} &\multicolumn{2}{c|}{R-Precision (LB Graph)} & \multirow{2}*{params (M)}\\
Graph& Image & K=50 & K=100 & K=50 & K=100& \\
\hline 
4 & 4& 65.7 &54.9 &76.1 &66.5 & 41.2/41.3 \\
4 & 6& 67.5 &57.3 &78.2 &69.4 & 45.4/45.5 \\
6 & 4& 67.2 & 56.5 & 78.3 & 70.4 & 45.4/45.5 \\
$\bm{6}$ & $\bm{6}$&  67.9 & 58.0 & 78.9 & 71.0 & 49.6/49.7 \\
6 & 8&  68.0 & 58.3 & 78.3 & 70.1  & 53.8/53.9 \\
8 & 6& 67.6 &57.1 &78.4 &69.7 & 53.8/53.9 \\
8 & 8& 67.1 &56.3 &78.0 &69.2 & 58.0/58.1 \\
\hline \hline
\end{tabular}
\end{adjustbox}
\caption{Ablation study for the Transformer layer number. Finally, we adopt a layer number of 6 for both the graph Transformer and the image Transformer.
For parameters, the left numbers are for location-free scene graphs, while the right numbers are for location-bounded scene graphs.
}
\label{tab:ablate_transformer}
\end{table}

\subsubsection{Impact of Batch Size}

\begin{figure}[tp]
\centering
\includegraphics[width=1\linewidth]{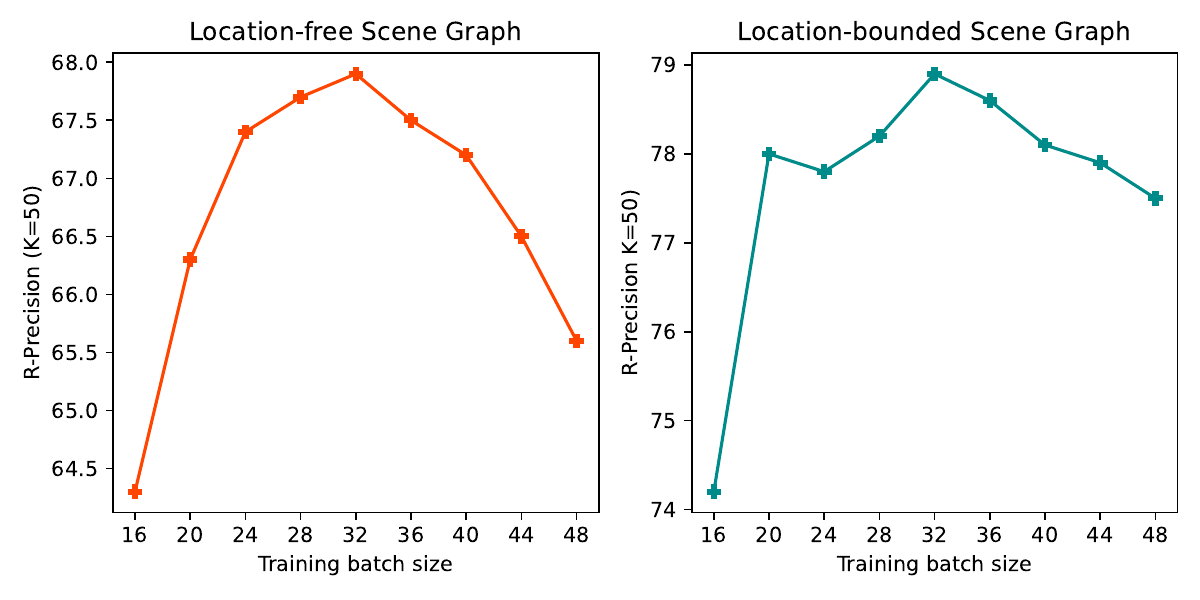}
\caption{We conduct experiments using different batch sizes to train our framework and finally adopt a batch size of 32.}
\label{fig:batch}
\end{figure}

Furthermore, we conduct ablative experiments with different batch sizes using our framework and evaluate SPAN based on R-Precision (K=50) as the evaluation metric. 
The results are illustrated in~Fig.~\ref{fig:batch}, which indicate that using batch sizes between 20 and 40 yields similar results for both location-free and location-bounded scene graphs.
The selection of a batch size that is too small or too large can lead to performance drops.
Too large a batch size is likely to result in similar graph-image pairs in the batch.
We finally use a batch size of 32 for training.

\subsubsection{Impact of Structural Encodings and Node Shuffle}

Moreover, we conduct ablative experiments to analyze the impact of structural encodings and node shuffle in graph serialization.
The results are demonstrated in Tab. \ref{tab:ablate_transformer}.
The Transformer outperforms the GCN for both location-free and location-bounded scene graphs with fewer parameters.
However, without structural encoding, the Transformer is unable to comprehend the graph structure, resulting in a significant performance drop.
In addition, we find node shuffle efficiently prevents biases resulting from the consistent assignment of node encodings to specific node representations, thus improving the overall performance.



\section{Conclusion}
 \label{sec:con}
In this paper, we propose a novel scene graph-image contrastive learning framework (SPAN) based on Transformers. 
In order to use the Transformer to reason about scene graph representations, we introduce a graph serialization approach that flattens a scene graph into a scene graph sequence with the structural encoding.
Our framework enables the quantitative measurement of similarity between scene graphs and images.
To address the limitations of existing metrics for scene graph generation, we propose a graph-oriented metric R-Precision based on SPAN.
The new metric indicates the model performance by evaluating the alignment between the generated scene graphs and the images, which strengthens the existing SGG evaluation. 
The new benchmarks are created for the Visual Genome and Open Images datasets.
Our graph Transformer can be further used as a generic scene graph encoder for other downstream tasks.
The source code and trained models are released publicly.

\ifCLASSOPTIONcompsoc
  \section*{Acknowledgments}
\else
  \section*{Acknowledgment}
\fi
This work has been supported by the Federal Ministry
of Education and Research (BMBF), under the
project LeibnizKILabor (grant no. 01DD20003), the Center for Digital Innovations (ZDIN), the Deutsche Forschungsgemeinschaft (DFG) under Germany’s Excellence Strategy within the Cluster of Excellence PhoenixD (EXC 2122), and EU HORIZON-CL42023-HUMAN-01-CNECT XTREME  (grant no.101136006).


 {
 \bibliographystyle{IEEEtran}
 \bibliography{egbib}
 }

\end{document}